\documentclass{article}
\usepackage{arxiv}

\usepackage{xspace}
\usepackage[utf8]{inputenc}
\usepackage[T1]{fontenc}
\usepackage[scaled=0.9]{helvet}
\DeclareMathAlphabet{\mathsf}{T1}{phv}{m}{n}
\usepackage{microtype}
\usepackage{graphicx}
\usepackage{amsmath,amssymb,amsthm}
\usepackage{cite}
\usepackage{booktabs}

\numberwithin{figure}{section}
\numberwithin{table}{section}
\numberwithin{equation}{section}

\usepackage[section]{algorithm}
\usepackage[noend]{algpseudocode}
\algrenewcommand{\algorithmiccomment}[1]{\unskip\nobreak\hfil\penalty50\hskip1em\hbox{}\nobreak\hfill{\color{black!60}$\triangleright$~#1}{\parfillskip=0pt\finalhyphendemerits=0\par}}
\newcommand{\floatsize}{\small}

\theoremstyle{definition}
\newtheorem{definition}{Definition}[section]
\newtheorem{example}{Example}[section]

\usepackage{xcolor}
\definecolor{RoyalBlue}{RGB}{65, 105, 225}
\definecolor{NaiveBarColor}{HTML}{C78283}
\definecolor{FullBarColor}{HTML}{495C83}
\PassOptionsToPackage{hyphens}{url}
\usepackage{hyperref}
\usepackage{xurl}
\usepackage{orcidlink}
\hypersetup{colorlinks=true,linkcolor=blue!75!black,citecolor=blue!75!black,urlcolor=blue!75!black}
\usepackage[most]{tcolorbox}
\tcbset{paperbox/.style={colback=white, colframe=black!35, boxrule=0.5pt, arc=1pt,
  left=6pt, right=6pt, top=5pt, bottom=5pt,
  colbacktitle=gray!10, coltitle=black, fonttitle=\bfseries, titlerule=0.5pt}}

\DeclareMathOperator*{\argmin}{arg\,min}

\usepackage{mathpartir}
\usepackage[normalem]{ulem}
\usepackage{enumitem}
\usepackage{mdframed}

\usepackage{multirow}
\usepackage{colortbl}
\usepackage{tabularx}
\usepackage{array}

\usepackage{pgfplots}
\usepackage{pgfplotstable}
\pgfplotsset{compat=1.17}
\usetikzlibrary{positioning}
\usepgfplotslibrary{statistics}

\newcommand{\tool}{\mbox{\textsf{MathEye}}\xspace}

\newcommand{\set}[1]{\{ #1 \}}

\algnewcommand\algorithmicforeach{\textbf{for each}}
\algdef{S}[FOR]{ForEach}[1]{\algorithmicforeach\ #1\ \algorithmicdo}

\newcommand{\reducedstrut}{\vrule width 0pt height .1\ht\strutbox depth .1\dp\strutbox\relax}
\newcommand{\icode}[1]{\begingroup\setlength{\fboxsep}{0pt}\colorbox{gray!20}{\reducedstrut{\sf #1}\/}\endgroup}
\newcommand{\irule}[1]{\begingroup\setlength{\fboxsep}{0pt}\colorbox{RoyalBlue!15}{\reducedstrut(\textsc{#1})\/}\endgroup}

\mdfsetup{nobreak=true}

\title{From Sorry to QED: Refining LLM Math Proofs\\ via Angelic Execution}

\renewcommand{\shorttitle}{From Sorry to QED: Refining LLM Math Proofs via Angelic Execution}

\date{}

\author{Yanju Chen\,\orcidlink{0000-0002-6494-3126}\textsuperscript{1}\thanks{Corresponding authors.}\quad
  Hanzhi Liu\,\orcidlink{0000-0002-7027-8302}\textsuperscript{2}\quad
  Ruizhe Qian\,\orcidlink{0009-0001-7983-2939}\textsuperscript{3}\quad
  Jiaming Shan\,\orcidlink{0009-0004-5239-9356}\textsuperscript{2}\\
  \bfseries
  Xu Yang\,\orcidlink{0000-0001-6239-1922}\textsuperscript{4}\footnotemark[1]\quad
  Yufei Ding\,\orcidlink{0000-0002-8716-5793}\textsuperscript{1}\quad
  Osbert Bastani\,\orcidlink{0000-0001-9990-7566}\textsuperscript{5}\quad
  Yu Feng\,\orcidlink{0000-0003-1000-1229}\textsuperscript{2}\footnotemark[1]\\[1.5ex]
  \small\textsuperscript{1}Department of Computer Science and Engineering, University of California, San Diego, La Jolla, CA 92093, USA\\
  \small\textsuperscript{2}Department of Computer Science, University of California, Santa Barbara, Santa Barbara, CA 93106, USA\\
  \small\textsuperscript{3}Department of Mathematics, Pennsylvania State University, University Park, PA 16802, USA\\
  \small\textsuperscript{4}Department of Mathematics, University of California, Santa Barbara, Santa Barbara, CA 93106, USA\\
  \small\textsuperscript{5}Department of Computer and Information Science, University of Pennsylvania, Philadelphia, PA 19104, USA\\[1ex]
  \small\ttfamily \{yanju, yufeiding\}@ucsd.edu\quad \{hanzhi, jiamingshan, xuyang, yufeng\}@ucsb.edu\\
  \small\ttfamily rbq5037@psu.edu\quad obastani@seas.upenn.edu
}

\hypersetup{
  pdftitle={From Sorry to QED: Refining LLM Math Proofs via Angelic Execution},
  pdfauthor={Yanju Chen, Hanzhi Liu, Ruizhe Qian, Jiaming Shan, Xu Yang, Yufei Ding, Osbert Bastani, Yu Feng},
  pdfkeywords={Interactive Theorem Proving, Large Language Models, Human-in-the-Loop, Formal Verification, Proof Synthesis},
}

\begin{document}

\maketitle
\setcounter{footnote}{0}

\begin{abstract}

Large language models (LLMs) can now generate Lean proof scripts that capture the high-level structure of many mathematical arguments. However, these scripts frequently fail on complex theorems: a single missing side condition or misapplied lemma can cause the prover to abort early and discard an otherwise viable proof outline. This paper presents \tool, a framework that converts imperfect LLM-generated proof scripts into fully verified Lean proofs by combining symbolic reasoning with minimal human interaction. At the core of \tool is \emph{angelic proof checking}, an optimistic execution model that replaces failing proof fragments with placeholders while preserving the overall structure of the argument. This optimistic check produces an \emph{angelic proof} that isolates the exact proof obligations that the LLM fails to discharge.

Because these unresolved obligations are often logically interdependent, asking a mathematician to address every placeholder directly is inefficient. To minimize human effort, \tool performs \emph{residual analysis} over a \emph{proof dependence graph} that captures symbol-level data flow in the angelic proof. By formulating obligation selection as a MaxSAT optimization problem, \tool selects a minimal-cost set of essential obligations whose resolution suffices to complete the entire proof. \tool presents these essential obligations to the mathematician in ordinary mathematical language and uses the mathematician's answers to automatically reconstruct the complete Lean proof. Experiments on 30 real-world benchmarks across analysis and algebra show that \tool solves 93.3\% of the problems while reducing the number of required user interactions by 55\%.

\end{abstract}


\section{Introduction}\label{sec:intro}

Formalizing mathematical arguments is critical for building reliable computer systems, including cryptographic reductions~\cite{coda}, numerical solvers~\cite{AppelK24}, verified compilers~\cite{compcert}, and distributed protocols~\cite{verdi}. When mechanized in proof assistants such as Lean~\cite{lean} or Coq~\cite{coqart}, mathematical proofs become machine-checkable artifacts that can be searched, refactored, and composed like software~\cite{avigad2018mechanizing}. Despite these benefits, constructing formal proofs remains expensive and time-consuming. Even routine proofs can require hours or days of expert effort, and relatively few mathematicians are fluent in modern proof assistants~\cite{hales2017formal}. As a result, the corpus of formalized mathematics continues to lag far behind the informal literature~\cite{wiedijk2008formal}.

Recent work on large language models (LLMs), including Minerva~\cite{minerva2022}, Llemma~\cite{llemma}, GPT-f~\cite{polu2022formal}, Thor~\cite{jiang2022thor}, and DeepSeek-Prover~\cite{deepseek2025prover}, has shown promise in automating formal proof construction. In particular, these models can translate natural-language mathematical problems into formal proof scripts in Lean or Coq~\cite{wu2022autoformalization} and drive interactive proof assistants tactic by tactic~\cite{yang2019learning}. However, while purely automated provers routinely succeed on short competition-style problems, they often struggle with longer and more complex theorems~\cite{deepseek2025prover,avigad2018mechanizing}. On such problems, a single missing side condition (e.g., continuity of an integrand, an inequality constraint, or a domain restriction) can cause a tactic to fail and lead the prover to reject an otherwise sound chain of reasoning, forcing proof search to restart from scratch. Moreover, many real-world proofs rely on conceptual decompositions and domain-specific insights that current automated provers cannot reliably infer end-to-end. Prior work in interactive program verification~\cite{isil,mayur} suggests an effective alternative: combining automated reasoning with targeted expert insight so that automation handles routine symbolic steps while humans intervene only where models struggle.

In this paper, we present \tool, a framework that transforms imperfect LLM-generated Lean proof drafts into fully verified proofs through minimal human interaction. Given a theorem statement, \tool first queries an off-the-shelf LLM to generate an initial Lean proof draft that captures the global structure of the intended argument. Rather than aborting at the first error, \tool checks this draft optimistically: whenever a tactic fails or leaves a precondition unsatisfied, \tool uses fault localization to replace the failing fragment with a placeholder (i.e., \icode{sorry}). This optimistic checking process produces an \emph{angelic proof}, a partially complete but structurally valid script that preserves the outline of the argument while isolating the unproven statements as \emph{residual goals}.

However, asking a mathematician to discharge every placeholder directly is impractical, because LLM-generated drafts often contain many redundant or logically dependent holes. To minimize human effort, \tool performs \emph{residual analysis} to identify a minimal subset of essential obligations whose resolution suffices to complete the entire proof. Specifically, \tool constructs a \emph{proof dependence graph} (PDG), a directed hypergraph in which nodes represent Lean symbols (such as hypotheses and intermediate lemmas) and hyperedges capture how proof fragments consume and produce these symbols. Using the PDG, \tool formulates obligation selection as a MaxSAT optimization problem over a cost model of human verification effort, finding a minimal-cost set of unproven statements from which all downstream obligations become derivable. \tool then presents these selected obligations to the mathematician in ordinary mathematical language rather than in Lean syntax. Once the mathematician confirms these claims, \tool automatically reconstructs the remaining proof and produces a complete, verified Lean development.

We have implemented \tool for Lean~4 and evaluated it on 30 real-world benchmarks across analysis and algebra, covering measure and integration theory, real and functional analysis, and group, ring, field, and polynomial theory. These benchmarks involve multi-step arguments with nontrivial intermediate lemmas that state-of-the-art LLM provers fail to complete end-to-end. Our evaluation shows that \tool solves 93.3\% (28 of 30) of the benchmarks through lightweight interaction with human participants, whereas the five automated systems we compare against complete at most 30.0\% (9 of 30) even when their results are combined. Furthermore, our ablation study demonstrates that residual analysis over the proof dependence graph reduces the total number of user queries by 55\% compared to querying all residual goals directly.

To summarize, this paper makes the following key contributions:
\begin{itemize}
    \item \textbf{Optimistic proof checking.}
    We introduce an optimistic, fault-localizing execution model that checks LLM-generated Lean scripts under angelic semantics, isolating failing proof fragments as symbolic placeholders while preserving the overall proof structure.

    \item \textbf{Proof dependence graph and residual analysis.}
    We develop a proof dependence graph to capture symbol-level data flow in angelic proofs, and formulate residual analysis as a MaxSAT optimization problem that selects a minimal-cost set of essential obligations.

    \item \textbf{Minimal-interaction refinement.}
    We design an interactive refinement workflow that queries the mathematician only on the essential obligations identified by residual analysis, in ordinary mathematical language, and automatically reconstructs the complete Lean proof.

    \item \textbf{Implementation and evaluation.}
    We implement \tool for Lean~4 and evaluate it on 30 real-world benchmarks across analysis and algebra. Our results show that \tool solves 93.3\% of these benchmarks while reducing the number of user queries by 55\%.
\end{itemize}

\section{Overview}\label{sec:overview}

In this section, we give a high-level overview of \tool using a motivating example that illustrates the gap between an informal mathematical argument and a machine-checked formal proof. Consider the following real analysis problem:

\begin{mdframed}[
    backgroundcolor=gray!10,
    linecolor=white,
    innertopmargin=0.5em,
    innerbottommargin=0.5em,
    ]\noindent
    \uline{\bf Problem:}
    Prove that the following inequality holds for all odd $C^1$ functions $f: [-1,1] \rightarrow \mathbb{R}$:
    \[ \int_{-1}^1 |f(x)|^2\,\mathrm{d}x \leq \int_{-1}^1 |f'(x)|^2\,\mathrm{d}x, \]
    where oddness means that $f(-x) = -f(x)$ for all $x \in [-1,1]$.
\end{mdframed}

\noindent On paper, a standard proof relates the magnitude of $f$ to that of its derivative $f'$ on the symmetric interval $[-1,1]$ in three main steps:
\begin{enumerate}
    \item \emph{Rewriting}. Because $f$ is odd, we have $f(0) = 0$. Combined with the $C^1$ assumption, the \emph{Fundamental Theorem of Calculus (FTC)} rewrites $f(x)$ in integral form:
    \[ f(x) = \int_0^x f'(t)\,\mathrm{d}t, \]
    reducing the problem to bounding an integral of $f'$.

    \item \emph{Bounding}. Applying the \emph{Cauchy--Schwarz inequality} on $[0,x]$ yields a pointwise bound on $|f(x)|^2$ for $x \in [0,1]$:
    \[ |f(x)|^2 \leq |x| \int_0^x |f'(t)|^2\,\mathrm{d}t \leq \int_0^1 |f'(t)|^2\,\mathrm{d}t. \]

    \item \emph{Symmetrizing}. Because $f$ is odd, both $|f|^2$ and $|f'|^2$ are even functions. Consequently, their integrals over $[-1, 1]$ are twice their integrals over $[0, 1]$, which lifts the pointwise bound on $[0,1]$ to the target $L^2$ inequality on $[-1,1]$.
\end{enumerate}
While this three-step argument is routine on paper, translating it into a fully verified Lean proof requires substantial effort. Each step relies on implicit regularity conditions, measure-theoretic justifications, and endpoint constraints that automated provers struggle to discharge.

\begin{figure}
    \centering
    \includegraphics[width=\linewidth]{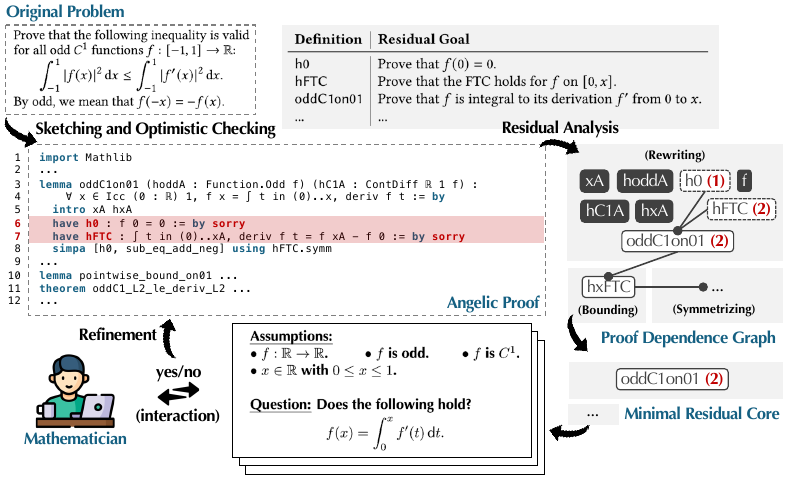}
    \caption{A motivating example demonstrating the workflow of \tool.}
    \label{fig:motivating}
\end{figure}

\paragraph{Limitations and challenges of existing approaches.}

For multi-step theorems of this kind, state-of-the-art LLM provers rarely succeed in a single pass. Even when a model generates a sound high-level proof outline, the resulting Lean script typically breaks down on low-level proof obligations. In our running example (\autoref{fig:motivating}), the LLM-generated sketch captures the three-step decomposition above (defining helper lemmas such as \icode{oddC1on01} at line~3 and \icode{pointwise\_bound\_on01} at line~10 before the main theorem \icode{oddC1\_L2\_le\_deriv\_L2} at line~11), but leaves multiple unresolved subgoals, or \emph{residual goals}, spread across the proof (top right of \autoref{fig:motivating}).

Specifically, consider the first helper lemma, \icode{oddC1on01} (lines 3--8 of \autoref{fig:motivating}), which formalizes the \emph{Rewriting} step on $[0, 1]$. Its proof fails on two intermediate subgoals in lines 6--7 that the LLM cannot discharge:
\begin{itemize}
    \item \icode{h0}: Prove that $f(0) = 0$.
    \item \icode{hFTC}: Prove that $\int_0^{x_A} f'(t)\,\mathrm{d}t = f(x_A) - f(0)$ for $x_A \in [0, 1]$.
\end{itemize}
Mathematically, both facts are elementary: \icode{h0} follows immediately from the oddness of $f$, and \icode{hFTC} is a direct application of the Fundamental Theorem of Calculus under the $C^1$ assumption. In Lean, however, proving \icode{hFTC} requires discharging a chain of library preconditions: the script must use the $C^1$ hypothesis (\icode{hC1A}) to establish that $f'$ is continuous, and then invoke additional Mathlib lemmas to show that $f'$ is measurable and interval-integrable on $[0, x_A]$. Because the LLM fails to supply these low-level justifications, Lean rejects \icode{oddC1on01} at lines 6--7, even though line~8 (\icode{simpa [h0, sub\_eq\_add\_neg] using hFTC.symm}) correctly derives the lemma's conclusion from \icode{h0} and \icode{hFTC}.

\paragraph{Our approach.}

\tool overcomes these limitations by combining automated symbolic analysis with targeted human insight. Rather than discarding a viable proof sketch when low-level tactics fail, \tool preserves the valid high-level structure and queries a mathematician only on a minimal set of essential obligations. Our methodology is built on three key ideas:
\begin{itemize}
    \item \emph{Optimistic Proof Checking.} \tool executes imperfect proof scripts under optimistic semantics: when a tactic or subproof fails, fault localization replaces the failing fragment with a \icode{sorry} placeholder so that Lean can continue checking the rest of the script. This produces an \emph{angelic proof} that validates the high-level deduction while collecting the unproven steps as \emph{residual goals}.

    \item \emph{External Certification.} Instead of requiring automated search to synthesize every low-level library precondition from scratch, \tool supports \emph{external certification}, allowing a mathematician to confirm intermediate claims that are mathematically routine but cumbersome to formalize in Lean.

    \item \emph{Dependency Analysis for Minimal Effort.} Because an angelic proof often contains many interdependent placeholders, \tool performs \emph{residual analysis} over a \emph{proof dependence graph} (PDG). By analyzing how unproven statements depend on one another, \tool extracts a \emph{minimal residual core}: a minimal-cost subset of obligations whose confirmation suffices to resolve all remaining goals.
\end{itemize}

\autoref{fig:motivating} illustrates how \tool applies these ideas to our running example. Starting from the original problem (top left), \tool queries an LLM for an initial proof draft and runs optimistic checking. Replacing the failed subproofs of \icode{h0} and \icode{hFTC} in lines 6--7 with \icode{sorry} allows Lean to accept line~8 and check the rest of the script. This produces the \emph{Angelic Proof} on the left of \autoref{fig:motivating}, together with the table of residual goals at the top right.

Rather than querying the mathematician on every residual goal in isolation, \tool constructs the \emph{Proof Dependence Graph} on the right of \autoref{fig:motivating} to capture symbol-level dependencies across the Rewriting, Bounding, and Symmetrizing stages. In this graph, solid black nodes (\icode{f}, \icode{hoddA}, \icode{hC1A}, \icode{xA}, \icode{hxA}) denote given hypotheses and parameters, dashed nodes (\icode{h0}, \icode{hFTC}) denote leaf \icode{sorry} obligations, and solid white nodes (\icode{oddC1on01}, \icode{hxFTC}) denote unproven lemmas derived from upstream obligations. Each candidate obligation is annotated in parentheses with its estimated verification cost. The bounding step \icode{hxFTC} depends on \icode{oddC1on01} in the Rewriting subgraph, which in turn depends on \icode{h0} (cost~1) and \icode{hFTC} (cost~2). Querying the leaf obligations \icode{h0} and \icode{hFTC} would require two separate interactions with a total cost of~3. By solving a MaxSAT optimization problem over the PDG, \tool determines that certifying the single lemma \icode{oddC1on01} (cost~2) resolves all downstream dependencies in this subgraph with only one query and a lower total cost.

Finally, during \emph{Refinement} (bottom of \autoref{fig:motivating}), \tool translates each obligation in the minimal residual core into a self-contained question in ordinary mathematical notation. For \icode{oddC1on01}, \tool lists the active assumptions ($f : \mathbb{R} \to \mathbb{R}$, $f$ is odd, $f$ is $C^1$, and $x \in \mathbb{R}$ with $0 \le x \le 1$) and asks the mathematician: ``Does the following hold? $f(x) = \int_0^x f'(t)\, \mathrm{d}t$.'' A single \emph{yes/no} response confirms this claim without exposing Lean syntax, allowing \tool to reconstruct the complete verified proof.

\paragraph{Framework overview.}

\begin{figure}
    \centering
    \includegraphics[width=0.85\linewidth]{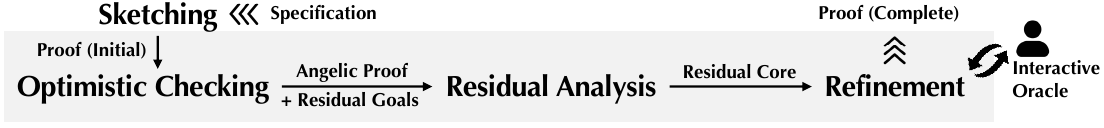}
    \caption{Framework overview of \tool.}
    \label{fig:framework}
\end{figure}

\autoref{fig:framework} summarizes the end-to-end architecture of \tool, which organizes proof refinement into four stages:
\begin{enumerate}
    \item \emph{Sketching.} Given a formal problem \emph{Specification}, \tool queries an LLM to decompose the theorem into intermediate lemmas and generate an initial Lean proof script (\emph{Proof (Initial)}).
    \item \emph{Optimistic Checking.} \tool executes the initial proof script in Lean, localizes failing tactic blocks, and replaces them with \icode{sorry} placeholders to produce an \emph{Angelic Proof} along with its set of \emph{Residual Goals} (\autoref{sec:refinement}).
    \item \emph{Residual Analysis.} From the angelic proof, \tool constructs a proof dependence graph, infers implicit mathematical relations, and solves a MaxSAT optimization problem to extract a minimal \emph{Residual Core} (\autoref{sec:analysis}).
    \item \emph{Refinement.} \tool presents the minimal residual core to the \emph{Interactive Oracle} (a mathematician) in ordinary mathematical language and uses the responses to reconstruct a complete, verified Lean proof (\emph{Proof (Complete)}, \autoref{sec:refinement}).
\end{enumerate}

\section{Preliminaries and Problem Formulation}\label{sec:problem}

In this section, we formalize proof states, proof script execution under optimistic semantics, and the residual goal resolution problem.

\begin{definition}[Proof State]
    A \emph{proof state} $\Gamma$ is a pair $(H, g)$, where $H = \{ h_1, \ldots, h_\ell \}$ is a finite set of \emph{hypotheses} (the local assumptions in scope) and $g$ is the \emph{goal} (the proposition to be established).
\end{definition}

In an interactive theorem prover such as Lean, a \emph{proof script} $P$ is a sequence of tactics that transforms an initial proof state by decomposing or discharging its goal. We model the execution of a proof script $P$ on a proof state $\Gamma$ using the execution judgment
\[ \Gamma \vdash P \Downarrow (\rho, Q), \]
where $\rho \in \{\top, \bot\}$ denotes the \emph{execution status} and $Q$ is a set of unresolved proof states collected during execution. Depending on $\rho$ and $Q$, this judgment captures three possible outcomes:

\begin{itemize}
    \item \emph{Complete proof} ($\rho = \top$ and $Q = \varnothing$): The script $P$ executes without error and discharges the goal $g$, leaving no remaining proof obligations.

    \item \emph{Hard failure} ($\rho = \bot$ and $Q = \varnothing$): Execution halts because a tactic raises an error (e.g., due to failed unification or an unsatisfied precondition) or the script terminates with an unhandled open goal.

    \item \emph{Angelic proof} ($\rho = \top$ and $Q = \{ \Gamma_1, \ldots, \Gamma_n \}$ with $n \ge 1$): The script $P$ is structurally valid but contains placeholders (\icode{sorry}) for one or more subproofs. Whenever the executor encounters a placeholder at a proof state $\Gamma_i$, it records $\Gamma_i$ in $Q$ and continues checking the remainder of $P$. The set $Q$ thus collects the \emph{residual goals} that remain to be justified.
\end{itemize}

\begin{figure}
    \centering
    \includegraphics[width=\linewidth]{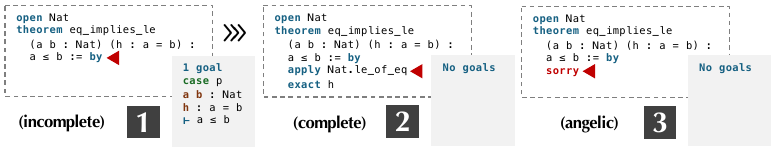}
    \caption{Three proof attempts for the theorem $a = b \Rightarrow a \le b$ in Lean and their execution outcomes: (1) an incomplete proof resulting in a hard failure, (2) a complete proof, and (3) an angelic proof.}
    \label{fig:ex-proof}
\end{figure}

\begin{example}[Proof Execution Outcomes]
    \autoref{fig:ex-proof} illustrates the three execution outcomes on a simple Lean theorem \icode{eq\_implies\_le}, which states that equality implies reflexive ordering on natural numbers ($a = b \Rightarrow a \leq b$). All three scripts start from the same initial proof state $\Gamma_0 = (H_0, g_0)$, where $H_0 = \{a, b : \text{Nat},\; h : a = b\}$ and $g_0 = (a \leq b)$:
    \begin{itemize}
        \item Case~1 shows an incomplete script $P_1$ with an empty tactic block, leaving $g_0$ unsolved without a placeholder. Execution halts with an unhandled open goal, yielding $\Gamma_0 \vdash P_1 \Downarrow (\bot, \varnothing)$.

        \item Case~2 shows the complete script $P_2 \equiv \icode{apply Nat.le\_of\_eq; exact h}$, which discharges $g_0$ and leaves no open goals, yielding $\Gamma_0 \vdash P_2 \Downarrow (\top, \varnothing)$.

        \item Case~3 shows the script $P_3 \equiv \icode{sorry}$. Upon encountering the \icode{sorry} placeholder, the executor records the current proof state $\Gamma_0$ as a residual goal and succeeds optimistically, yielding $\Gamma_0 \vdash P_3 \Downarrow (\top, \{ \Gamma_0 \})$.
    \end{itemize}
\end{example}

\paragraph{Problem formulation.}

Intuitively, the execution judgment $\Gamma \vdash P \Downarrow (\rho, Q)$ separates scripts that fail outright ($\rho = \bot$) from angelic proofs ($\rho = \top, Q \ne \varnothing$) that preserve a valid high-level deduction modulo unresolved placeholders. Given an angelic proof, completing the formal verification reduces to discharging each residual goal in $Q$, which we formalize as follows.

\begin{definition}[The Residual Goal Resolution Problem]
A \emph{resolution} for a proof state $\Gamma = (H, g)$ is either:
\begin{enumerate}
    \item a \emph{proof witness} (e.g., a Lean term or tactic sequence) that establishes $g$ from the hypotheses $H$, or
    \item a \emph{certified assumption} supplied by a human oracle that asserts $g$ holds under~$H$ and may be used as an auxiliary fact in subsequent checking.
\end{enumerate}

Given an angelic proof $P'$ for an initial proof state $\Gamma_0 = (H_0, g_0)$ such that
\[
    \Gamma_0 \vdash P' \Downarrow (\top, Q)
    \qquad\text{with}\qquad 
    Q \neq \varnothing,
\]
the \emph{residual goal resolution problem} is to compute a set of resolutions
$\mathcal{R} = \{ r_1,\ldots,r_n \}$ such that for every residual proof state
$\Gamma_i = (H_i, g_i) \in Q$, the corresponding resolution $r_i \in \mathcal{R}$
provides a valid justification for discharging $g_i$ under $H_i$.
\end{definition}

In the following sections, we show how \tool solves this problem while minimizing the number and difficulty of certified assumptions requested from the mathematician.

\section{Minimal-Interaction Refinement Algorithm}\label{sec:refinement}

In this section, we present our iterative refinement algorithm for converting a formal problem specification into a verified Lean proof. We first describe the top-level \textsc{Refine} procedure (\autoref{subsec:algorithm}) and then detail the \textsc{Check} procedure (\autoref{subsec:checking}), which constructs angelic proofs via structural fault localization. We defer the details of residual analysis (\textsc{Analyze}) to \autoref{sec:analysis}.

\subsection{Top-Level Algorithm}\label{subsec:algorithm}

\begin{algorithm}[t]\floatsize
    \begin{algorithmic}[1]
        \Procedure{Refine}{$\phi$}
            \State \textbf{input:} problem specification $\phi$ with initial proof state $\Gamma_0$; cost model $C$
            \State \textbf{output:} refined proof $P$ in formal language or $\bot$
            \State $P_0 \gets \textsc{Sketch}(\phi)$ \Comment{generate an initial proof}
            \State $P \gets \textsc{Check}(P_0)$ \Comment{obtain an angelic proof via optimistic checking}
            \While {$\Gamma_0 \vdash P \Downarrow (\top,Q)$} \Comment{repeat the refinement process unless the proof fails}
                \State {\bf if} $Q = \varnothing$ {\bf then} {\bf return} $P$ \Comment{no more residual goals to resolve; proof completes}
                \State $K \gets \textsc{Analyze}(P; C)$ \Comment{compute a minimal residual core}
                \State {\bf for} {\bf each} $v \in K$ {\bf do} $P \gets \textsc{Update}(P, v)$ \Comment{refine the proof with human oracle}
            \EndWhile
            \State {\bf return} $\bot$
        \EndProcedure
    \end{algorithmic}
    \caption{Minimal-Interaction Refinement}
    \label{alg:refine}
\end{algorithm} 

\autoref{alg:refine} summarizes our top-level refinement procedure, \textsc{Refine}, which takes as input a formal problem specification $\phi$ (with initial proof state $\Gamma_0$) and a cost model $C$ (\autoref{subsec:mrc}), and returns either a verified Lean proof $P$ or $\bot$ indicating failure. At a high level, \textsc{Refine} first constructs an initial angelic proof from an LLM-generated draft (lines 4--5) and then iteratively resolves the remaining obligations through targeted queries to the mathematician (lines 6--9).

In more detail, \textsc{Refine} begins by invoking \textsc{Sketch} (line 4), which queries an LLM to decompose $\phi$ into intermediate lemmas and emit an initial proof script $P_0$. Because $P_0$ typically contains failing tactics or unmet library preconditions, \textsc{Refine} passes $P_0$ to \textsc{Check} (line 5), which localizes the failing fragments and replaces them with \icode{sorry} placeholders to produce an initial angelic proof $P$.

The algorithm then enters the main refinement loop (lines 6--9). In each iteration, it evaluates the current script $P$ on $\Gamma_0$ under the execution judgment $\Gamma_0 \vdash P \Downarrow (\rho, Q)$ from \autoref{sec:problem}. If $P$ executes without error and leaves no residual goals ($Q = \varnothing$), then $P$ is a complete proof and the algorithm returns $P$ (line 7). Otherwise, \textsc{Refine} invokes \textsc{Analyze} (line 8) to compute a minimal residual core $K$ from $P$ under the cost model $C$. For each obligation $v \in K$, the \textsc{Update} procedure (line 9) translates the proof state at $v$ into ordinary mathematical notation and asks the mathematician whether the goal holds under its assumptions. If the mathematician confirms $v$, \textsc{Update} records $v$ as a certified assumption and incorporates it into $P$. If $P$ can no longer be executed optimistically ($\rho = \bot$), the loop terminates and the algorithm returns $\bot$ (line 10).

\subsection{Optimistic Checking}\label{subsec:checking}

Given a raw LLM-generated proof script $P_0$, the goal of the \textsc{Check} procedure (line 5 of \autoref{alg:refine}) is to transform $P_0$ into a structurally valid angelic proof $P$ satisfying $\Gamma_0 \vdash P \Downarrow (\top, Q)$. A key challenge in designing \textsc{Check} is choosing the right granularity at which to isolate errors. Replacing an entire theorem body with \icode{sorry} would trivially eliminate all errors, but would discard the LLM's decomposition entirely. Conversely, replacing only the single failing tactic can be insufficient, because the source of the error may lie at a higher level of the proof.

To preserve as much valid proof structure as possible while ensuring that the resulting script type-checks, \textsc{Check} performs bottom-up \emph{structural backtracking} over the abstract syntax tree (AST) of the proof script. After a lightweight syntactic normalization pass using standard parser error recovery, \textsc{Check} evaluates the script in Lean to identify semantic proof errors (i.e., tactic failures or undischarged subgoals). Whenever Lean reports an error at a tactic node, \textsc{Check} locates the enclosing tactic or subproof block in the AST, replaces that block with a \icode{sorry} placeholder, and re-evaluates the modified script. If the error is resolved, \textsc{Check} retains the replacement and continues checking the rest of the script. If the fault persists, indicating that the error's source, perhaps a missing hypothesis, lies at a higher level of the proof, \textsc{Check} ascends to the parent AST node and replaces the larger enclosing block with \icode{sorry}. This bottom-up backtracking repeats until all semantic faults are isolated within \icode{sorry} placeholders, yielding an angelic proof $P$ whose placeholders define the residual goals $Q$.

\begin{figure}[t]
    \centering
    \includegraphics[width=\linewidth]{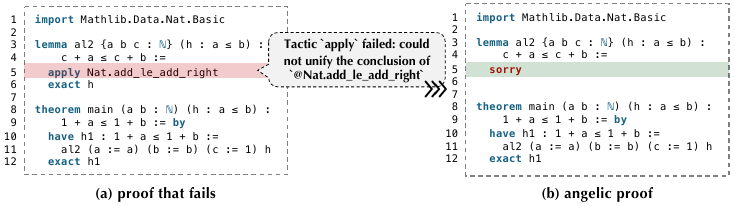}
    \caption{Fault localization and generation of an angelic proof via structural backtracking. (a) Initial incomplete script with a semantic error caused by a misapplied tactic. (b) Structurally valid angelic proof after backtracking replaces the failing proof block with a \icode{sorry} placeholder.}
    \label{fig:ex-angfix}
\end{figure}

\begin{example}[Fault Localization and Generation of Angelic Proof]
    Consider the initial proof script $P_0$ in \autoref{fig:ex-angfix}(a), which states a helper lemma \icode{al2} before proving \icode{theorem main}. At line~5, the proof of \icode{al2} fails because the tactic \icode{apply Nat.add\_le\_add\_right} cannot unify with the goal $c + a \le c + b$. When \textsc{Check} encounters this error, it backtracks to the enclosing proof block of \icode{al2} (lines 5--6) and replaces it with a single \icode{sorry} placeholder (line~5 of \autoref{fig:ex-angfix}(b)). With the type of \icode{al2} satisfied optimistically, Lean proceeds to check \icode{theorem main} (lines 8--12), which instantiates \icode{al2} to derive \icode{h1} and closes the main goal. The resulting script in \autoref{fig:ex-angfix}(b) is a valid angelic proof with a single residual goal at \icode{al2}.
\end{example}

\section{Residual Analysis}\label{sec:analysis}

Given an angelic proof $P$ with residual goals $Q$, querying the mathematician on every goal in $Q$ directly is often inefficient. In multi-step proofs, residual goals are rarely independent: some goals are immediate consequences of others, some are low-level preconditions of a single lemma, and others become redundant once a higher-level claim is established. The goal of our residual analysis procedure (\textsc{Analyze}, summarized in \autoref{alg:analyze}) is to capture these dependencies and extract a \emph{minimal residual core} $K$, a minimal-cost subset of obligations whose justification suffices to complete the entire proof.

\begin{algorithm}[t]\floatsize
    \begin{algorithmic}[1]
        \Procedure{Analyze}{$P$; $C$}
            \State \textbf{input:} proof $P$
            \State \textbf{output:} minimal residual core $K$ (parameterized by cost model $C$) or $\bot$
            \State $G = (V, E) \gets {\sf buildPDG}(P)$ \Comment{build proof dependence graph from given proof}
            \State $G' = (V, E')$, where $E' \gets E \cup \set{e\ |\ P,V,E \models e}$ \Comment{infer paths via math relations}
            \State $\Psi \gets \textsc{Encode}(G')$ \Comment{encode the graph into constraints}
            \State $m \gets {\sf minimize}(C; \Psi)$ \Comment{search for a model that minimizes the cost}
            \State {\bf if} ${\sf unsat}(m)$ {\bf then} {\bf return} $\bot$ \Comment{no model exists}
            \State {\bf else} {\bf return} $K$, where $K \gets {\sf choose}(V, m)$ \Comment{choose the corresponding nodes and return}
        \EndProcedure
    \end{algorithmic}
\caption{Residual Analysis}
\label{alg:analyze}
\end{algorithm} 

As shown in \autoref{alg:analyze}, \textsc{Analyze} takes as input an angelic proof $P$ and a cost model $C$, and computes the minimal residual core $K$ in three stages:
\begin{itemize}
    \item \emph{Proof Dependence Graph Construction (\autoref{subsec:pdg}):} First, ${\sf buildPDG}(P)$ constructs an explicit \emph{proof dependence graph} $G = (V, E)$ (line 4), a directed hypergraph that records symbol-level data flow among hypotheses, intermediate lemmas, and goals in $P$.

    \item \emph{Relational Inference (\autoref{subsec:mathrels}):} Because $P$ is derived from an incomplete draft, its explicit PDG can be disconnected when intermediate lemmas are stated but not syntactically referenced before a \icode{sorry} placeholder. \textsc{Analyze} therefore augments $G$ with implicit dependency hyperedges inferred via the relational judgment $P, V, E \models e$, producing the enriched graph $G' = (V, E')$ (line 5).

    \item \emph{Minimal Core Optimization (\autoref{subsec:mrc}):} Finally, \textsc{Encode} translates $G'$ into a set of hard propositional constraints $\Psi$ characterizing valid residual cores (line 6), and ${\sf minimize}(C; \Psi)$ solves a MaxSAT problem to find a satisfying assignment $m$ that minimizes the human effort cost $C$ (line 7). If $\Psi$ is unsatisfiable, \textsc{Analyze} returns $\bot$ (line 8); otherwise, it returns the subset of nodes $K \subseteq V$ selected by the optimal model $m$ (line 9).
\end{itemize}

\subsection{Proof Dependence Graph}\label{subsec:pdg}

Because a single proof step typically combines multiple assumptions and lemmas to establish a new fact, a standard directed graph cannot distinguish between conjunctive premises (all of which are required) and alternative proof paths. We therefore model the symbol-level data flow of a proof script $P$ as a directed hypergraph, defined as follows.

\begin{definition}[Proof Dependence Graph]
A \emph{proof dependence graph} (PDG) for a Lean proof script $P$ is a directed dependence hypergraph
\[ G = (V, E). \]
Each node $v \in V$ corresponds to a proof symbol introduced in $P$, including initial hypotheses, local parameters, and declarations created by commands such as \icode{have}, \icode{lemma}, \icode{theorem}, or \icode{let}. Each hyperedge
\[ e = ({\sf src}(e), {\sf dst}(e)) \in E \]
corresponds to a contiguous proof fragment whose body derives one or more target symbols ${\sf dst}(e) \subseteq V$ from the set of referenced source symbols ${\sf src}(e) \subseteq V$.
\end{definition}

\begin{figure}
    \centering
    \includegraphics[width=\linewidth]{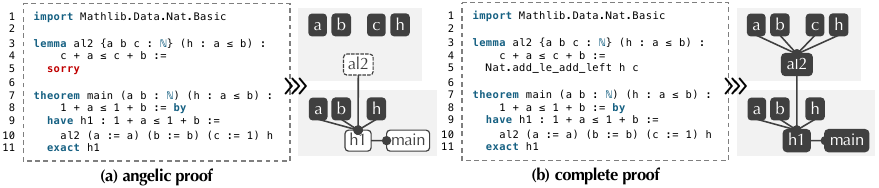}
    \caption{Angelic (a) and complete (b) versions of an example proof and their proof dependence graphs.}
    \label{fig:ex-pdg}
\end{figure}

\begin{example}[Proof Dependence Graphs for Complete Proofs]
    Consider the complete proof in \autoref{fig:ex-pdg}(b). The first proof fragment defines the lemma \icode{al2} by invoking \icode{Nat.add\_le\_add\_left} on hypothesis \icode{h} and parameters \icode{a}, \icode{b}, and \icode{c}. This fragment induces the hyperedge
    \[ \set{\icode{a}, \icode{b}, \icode{c}, \icode{h}} \ \rightarrow\ \set{\icode{al2}}. \]
    Inside \icode{theorem main}, the \icode{have} fragment defines the intermediate symbol \icode{h1} by applying \icode{al2} to local hypotheses $\set{\icode{a}, \icode{b}, \icode{h}}$, inducing the hyperedge
    \[ \set{\icode{a}, \icode{b}, \icode{h}, \icode{al2}} \ \rightarrow\ \{\icode{h1}\}. \]
    Finally, \icode{exact h1} establishes \icode{main} from \icode{h1}, inducing the hyperedge
    \[ \set{\icode{h1}} \ \rightarrow\ \set{\icode{main}}. \]
\end{example}

This construction extends naturally to angelic proofs, as shown in \autoref{fig:ex-pdg}(a). Because the body of \icode{al2} is replaced by \icode{sorry}, the fragment references no premises, so the incoming hyperedge from $\set{\icode{a}, \icode{b}, \icode{c}, \icode{h}}$ to \icode{al2} is absent. This missing incoming hyperedge identifies \icode{al2} as an unresolved residual goal.

\paragraph{Graphical convention.}

Throughout the paper, we use solid black nodes (e.g., \icode{a}, \icode{b}, \icode{h}) to denote axiomatized or proven symbols, and white nodes to denote unproven symbols that transitively depend on a residual goal. A dashed border (such as on \icode{al2} in \autoref{fig:ex-pdg}(a)) marks a node whose own proof fragment is \icode{sorry}.

\subsection{Inference of Math Relations}\label{subsec:mathrels}

\begin{figure}
    \centering
    \includegraphics[width=\linewidth]{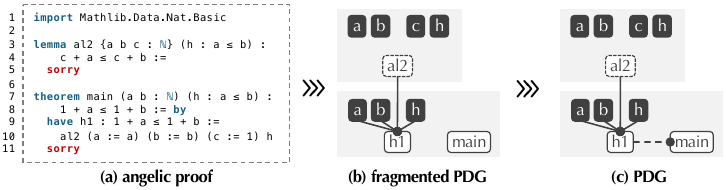}
    \caption{An angelic proof that produces a fragmented PDG, which is fixed by inference of math relations.}
    \label{fig:ex-mrels}
\end{figure}

Because ${\sf buildPDG}(P)$ only extracts dependencies that appear syntactically in the script $P$, the resulting graph $G = (V, E)$ can be fragmented when a proof block ends prematurely with \icode{sorry}. \autoref{fig:ex-mrels}(a) shows a typical case: the LLM states a useful helper lemma \icode{al2} and derives an intermediate fact \icode{h1} from it (line 10), but fails to close the main goal and leaves \icode{sorry} at line 11. As shown in \autoref{fig:ex-mrels}(b), the explicit PDG captures the hyperedge into \icode{h1}, but leaves \icode{main} disconnected from \icode{h1}. Optimizing over this fragmented graph directly would select \icode{main} itself as the obligation to query, ignoring the decomposition already established by \icode{al2} and \icode{h1}.

To bridge such gaps, \textsc{Analyze} infers implicit dependencies by inspecting the mathematical types of the symbols in $V$. We formalize this inference using a library of rules under the judgment
\[ P, V, E \models e, \]
which states that given the proof context $P$, node set $V$, and explicit hyperedges $E$, an implicit hyperedge $e$ can be inferred.

\begin{example}[Resolving the Fragmented PDG Problem]
    In \autoref{fig:ex-mrels}, a syntactic instantiation rule inspects the disconnected \icode{main} node and the symbols in scope. Because \icode{main}'s goal ($1 + a \le 1 + b$) matches the proposition of \icode{h1}, the rule infers the implicit edge $e = (\{\icode{h1}\}, \{\icode{main}\})$ shown in \autoref{fig:ex-mrels}(c). Adding this edge restores the dependency chain $\icode{al2} \rightarrow \icode{h1} \rightarrow \icode{main}$, allowing the optimizer to identify \icode{al2} as the upstream root obligation.
\end{example}

\begin{figure}[t]
    \small
    \mprset{andskip=2.5em minus 0.5em}
    \centering
    \begin{mathpar}[\lineskiplimit=0.7em\lineskip=0.7em plus 0.2em]
        \inferrule{
            f \in \mathbb{F} \\
            I_1, I_2 \in \mathbb{I} \\
            I_2 \subseteq I_1,\ I_2 \text{ compact} \\\\
            h_2: \oplus(f, I_2) \\
            h_1: {\sf continuous}(f, I_1) \\\\
            \oplus \in \{ {\sf bounded}, {\sf riemann\_integrable}, \dots \}
        }{
            P, V, E \models (\set{h_1}, \set{h_2})
        }\textsc{ (Continuity)}
        \and
        \inferrule{
            f : \mathbb{R} \to \mathbb{R}\\
            a, b \in \mathbb{R} \\\\
            h_1: a \le b\\
            h_2: f \in C^1(\mathbb{R}) \\\\
            h_3: \textstyle \int_a^b f'(t)\,dt = f(b) - f(a)
        }{
            P, V, E \models (\set{h_1, h_2}, \set{h_3})
        }\textsc{ (FTC)}
        \\
        \inferrule{
            a, b, c, d \in \mathbb{R} \\\\
            h_1: a < b \\
            h_2: c = 2a \\\\
            h_3: d = 2b \\
            h_4: c < d
        }{
            P, V, E \models (\set{h_1, h_2, h_3}, \set{h_4})
        }\textsc{ (Algebra)}
        \and
        \inferrule{
            A, B \in {\sf Set} \\
            f : A \to B \\
            h_1: {\sf injective}(f) \\\\
            h_2: |A| = \infty \\
            h_3: |B| = \infty
        }{
            P, V, E \models (\set{h_1, h_2}, \set{h_3})
        }\textsc{ (InfImage)}
        \\
        \inferrule{
            X \in {\sf Set} \\
            X \neq \varnothing \\
            {\le} \in {\sf PartialOrder}(X) \\\\
            h_2: |X| = \infty \\\\
            h_1: \forall x \in X.\ \exists z \in X.\ x \le z \land x \neq z
        }{
            P, V, E \models (\set{h_1}, \set{h_2})
        }\textsc{ (NoMaxElem)}
        \and
        \inferrule{
            \alpha \in {\sf Prop} \\\\
            h_1: \alpha \to {\sf False} \\
            h_2: \neg \alpha
        }{
            P, V, E \models (\set{h_1}, \set{h_2})
        }\textsc{ (NegIntro)}
        \\
        \inferrule{
            f: \mathbb{R} \to \mathbb{R} \\
            a, b \in \mathbb{R} \\
            h_1: a \le b \\\\
            h_2: {\sf riemann\_integrable}(f, [a,b]) \\\\
            h_3: \textstyle \int_{\mathbb{R}} f(x)\mathbf{1}_{[a,b]}(x)\,d\mu(x) = \textstyle \int_a^b f(x)\,dx
        }{
            P, V, E \models (\set{h_1, h_2}, \set{h_3})
        }\textsc{ (RL-Agree)}
        \and
        \inferrule{
            A, B \in {\sf Set} \\
            h_1: {\sf countable}(A) \\\\
            h_2: \exists f : A \to B,\; {\sf bijective}(f) \\\\
            h_3: {\sf countable}(B)
        }{
            P, V, E \models (\set{h_1, h_2}, \set{h_3})
        }\textsc{ (CtbleBij)}
        \\
        \inferrule{
            f: [0,1] \to \mathbb{R} \\
            h_1: {\sf lebesgue\_integrable}(f, [0,1]) \\\\
            h_2: \forall \varepsilon > 0\, \exists \tilde f : [0,1] \to \mathbb{R}, \\
            {\sf continuous}(\tilde f, [0,1]) \land \|f-\tilde f\|_{L^1} < \varepsilon
        }{
            P, V, E \models (\set{h_1}, \set{h_2})
        }\textsc{ ($L^1$-Dense)}
    \end{mathpar}
    \caption{Representative relational inference rules for residual analysis. Here $f \in \mathbb{F}$ and $I \in \mathbb{I}$ abbreviate that $f$ is a function and $I$ an interval in the proof context.}
    \label{fig:mrules}
\end{figure}

\autoref{fig:mrules} presents a representative subset of our relational inference rules. These rules capture standard mathematical implications that are routine in paper proofs, but often remain unconnected in LLM-generated Lean scripts when intermediate tactics fail. Specifically, our rule library spans three main domains:
\begin{itemize}
    \item \emph{Real Analysis.} Rules such as \irule{FTC}, \irule{RL-Agree}, and \irule{$L^1$-Dense} link analytical regularity conditions (such as $f \in C^1(\mathbb{R})$, Riemann integrability, or Lebesgue integrability) to integral identities and density properties. Similarly, \irule{Continuity} uses continuity on an interval $I_1$ to infer boundedness or integrability on any compact subinterval $I_2 \subseteq I_1$.

    \item \emph{Set Theory and Cardinality.} Rules such as \irule{InfImage}, \irule{NoMaxElem}, and \irule{CtbleBij} capture how cardinality properties propagate across mappings. For example, if a domain $A$ is infinite ($|A| = \infty$) or countable (${\sf countable}(A)$), an injective or bijective map $f : A \to B$ propagates the corresponding cardinality property to $B$.

    \item \emph{Algebra and Propositional Logic.} Rules such as \irule{Algebra} and \irule{NegIntro} capture linear inequality scaling and propositional negation introduction (inferring $\neg \alpha$ from $\alpha \to {\sf False}$), bridging local deductive steps when tactics fail.
\end{itemize}

Augmenting the explicit PDG with hyperedges derived from these rules recovers the intended dependency structure of the proof prior to optimization.

\subsection{Minimizing Residual Cores}\label{subsec:mrc}

Given the augmented proof dependence graph $G' = (V, E')$, our goal is to select a set of obligations $K$ whose confirmation by the mathematician suffices to establish the final theorem at minimal cost. Let $H \subseteq V$ denote the set of initial hypotheses and given definitions, and let $v_t \in V {\setminus} H$ denote the target theorem node.

We say that a node $v \in V$ is \emph{provable} from a seed set $W \subseteq V$ if $v \in W$, or if there exists a hyperedge $e \in E'$ such that $v \in {\sf dst}(e)$ and every source node $u \in {\sf src}(e)$ is provable from $W$. We refer to this recursive closure under hyperedges in $E'$ as \emph{proof propagation}. A subset of non-hypothesis nodes $K \subseteq V {\setminus} H$ is a \emph{valid residual core} if the target theorem $v_t$ is provable from $H \cup K$ via proof propagation.

Because any node provable from $H$ alone is already established by the valid fragments of the angelic proof, we restrict our search space to the set of \emph{unproven} nodes $U \subseteq V {\setminus} H$ that cannot be derived from $H$ alone. In general, $G'$ admits multiple valid cores $K \subseteq U$: for example, the set of leaf \icode{sorry} nodes (corresponding to the residual goals in $Q$) is a valid core, as is the singleton $\{v_t\}$. To select a valid core that minimizes human effort, we formulate minimal residual core selection as a constrained optimization problem solved via MaxSAT:
\[
K_\diamond = \argmin_{K \subseteq U}\; C(K; G, P) \quad \text{subject to } \Psi, \text{ where } \Psi = \textsc{Encode}(G').
\]
Here, the hard constraints $\Psi = \textsc{Encode}(G')$ characterize all valid residual cores under proof propagation, and the cost model $C(K; G, P)$ defines the soft objective measuring human verification effort.

\paragraph{Encoding.}

The \textsc{Encode} procedure translates $G' = (V, E')$ into a propositional formula $\Psi$ over two sets of Boolean variables:
\begin{itemize}
    \item For each node $v \in V$, a \emph{state variable} $x_v$, where $x_v = {\sf true}$ indicates that symbol $v$ is proven.
    \item For each non-hypothesis node $v \in V {\setminus} H$, a \emph{choice variable} $y_v$, where $y_v = {\sf true}$ indicates that $v$ is selected into the residual core $K$.
\end{itemize}
A satisfying assignment defines the residual core $K = \{v \in V {\setminus} H \mid y_v = {\sf true}\}$. The hard constraints $\Psi$ consist of five types of clauses:
\begin{enumerate}
    \item (Hypotheses) For every initial hypothesis $v \in H$, we add the unit clause $(x_v)$, asserting that given hypotheses are unconditionally proven.
    \item (Human Justification) For every $v \in V {\setminus} H$, we add the clause $(\neg y_v \lor x_v)$, asserting that if the oracle justifies $v$ ($y_v = {\sf true}$), then $v$ becomes proven ($x_v = {\sf true}$) regardless of its upstream dependencies.
    \item (Proof Propagation) For every hyperedge $e \in E'$, we add the implication
    \[ \left( \bigwedge_{u \in {\sf src}(e)} x_u \right) \implies \left( \bigwedge_{v \in {\sf dst}(e)} x_v \right), \]
    enforcing that if all source nodes of $e$ are proven, all destination nodes of $e$ are also proven.
    \item (Support) For every non-hypothesis node $v \in V {\setminus} H$, we add the converse implication
    \[ x_v \implies y_v \lor \bigvee_{e \in E',\, v \in {\sf dst}(e)} \ \bigwedge_{u \in {\sf src}(e)} x_u, \]
    ensuring that a node counts as proven only if the mathematician justifies it or some incoming hyperedge derives it from proven source nodes. Without this clause, setting every $x_v$ to ${\sf true}$ would trivially satisfy the remaining constraints with an empty core.
    \item (Target Resolution) Finally, we add the unit clause $(x_{v_t})$, requiring the target theorem $v_t$ to be proven.
\end{enumerate}

When the hypergraph is acyclic (as is the case when hyperedges only connect earlier declarations to later ones), the proven nodes of a satisfying assignment coincide with the nodes provable from $H \cup K$. Every satisfying assignment thus yields a valid residual core $K$, while nodes already provable from $H$ can be assigned $y_v = {\sf false}$ at zero cost.

\paragraph{Cost model.}

To distinguish among valid residual cores, our cost model $C$ balances the difficulty of evaluating individual obligations against the total number of user queries. For any non-empty candidate core $K \subseteq U$ in $G=(V,E)$, we define:
\[
C(K; G,P) = \frac{1}{|K|}\sum_{v \in K} (\lambda_{\sf sem} \delta(v; G, P) + \lambda_{\sf str} \sigma(v; G, P)) + \lambda_{\sf size} \frac{|K|}{|V|}.
\]
Here, the first term measures the average verification difficulty of the obligations in $K$: $\delta(v;G,P) \in [0,1]$ is an LLM-estimated \emph{semantic difficulty} score normalized between~$0$ (for given premises) and~$1$ (for the full theorem goal), while $\sigma(v;G,P)$ is a deterministic \emph{structural complexity} measure based on the size of the Lean proof state at~$v$, which penalizes large or cluttered contexts to account for goals that are mathematically simple but syntactically verbose. The weights $\lambda_{\sf sem}$ and $\lambda_{\sf str}$ balance semantic difficulty against syntactic state complexity. The second term, weighted by $\lambda_{\sf size} \ge 0$, penalizes the normalized core size $|K| / |V|$, biasing the solver toward smaller cores that require fewer interactions. Minimizing $C(K;G,P)$ subject to~$\Psi$ thus selects a compact set of obligations that are individually easy to assess and collectively sufficient to complete the proof.

\section{Implementation}\label{sec:impl}

We have implemented \tool in a toolchain that combines a Python back-end with a front-end built on Lean~4~\cite{lean} and Mathlib~\cite{mathlib2020}, with an emphasis on usability and reproducibility. For the initial decomposition phase (\textsc{Sketch}), \tool queries ChatGPT~5.1 Extended Thinking~\cite{openai_chatgpt} to partition each benchmark problem into Lean-typed subgoals. The remaining stages of the pipeline, including optimistic checking, residual analysis, and interactive refinement, are orchestrated in Python and rely on Lean for proof checking and final validation.

During optimistic checking (\textsc{Check}), the front-end runs the generated proof draft through Lean~4 and Mathlib and replaces failing proof fragments with \icode{sorry} placeholders until the script stabilizes as an angelic proof. The same front-end extracts the proof dependence graph, after which the analysis pipeline performs propagation, adds the relational hyperedges inferred by the rules from \autoref{subsec:mathrels}, and encodes the hard constraints $\Psi$ and the cost objective $C$ into the Z3 SMT solver via its \texttt{Optimize} module. Difficulty scores are used only to rank candidate residual cores, and the completed proof is always rechecked by Lean.

\section{Evaluation}\label{sec:eval}

In this section, we evaluate \tool through experiments designed to answer the following research questions:
\begin{itemize}
\item {\bf RQ1 (Overall Effectiveness):} Can \tool prove challenging real-world mathematical theorems that state-of-the-art automated provers fail to complete?
\item {\bf RQ2 (User Effort \& Interaction Cost):} How much human effort, measured in query count and response time, is required to complete proofs in \tool?
\item {\bf RQ3 (Ablation Study):} How much does residual analysis reduce the number of user interactions compared to querying all residual goals directly?
\end{itemize}

\subsection{Benchmarks and Selection Criteria}
To evaluate \tool, we curated a suite of 30 advanced mathematical benchmarks in Lean~4. While existing automated provers have been heavily optimized for competition-style datasets such as \textsc{PutnamBench}~\cite{putnambench}, datasets that existing models have often been trained and tuned on, they still struggle to formalize graduate-level mathematics. We selected our benchmarks to reflect three structural challenges characteristic of this setting:
\begin{itemize}
    \item \textit{Multi-Step Reasoning:} Each theorem requires a multi-step proof (typically spanning 7 to 10 major logical transitions) rather than shallow tactic search, requiring the prover to maintain a complex proof state across intermediate steps.
    \item \textit{Strict Side Conditions:} The proofs require formulating intermediate lemmas and discharging implicit preconditions, such as measure-theoretic regularity conditions or algebraic domain constraints.
    \item \textit{End-to-End Theorem Proving:} Each benchmark is a complete theorem statement rather than an isolated subgoal.
\end{itemize}

Following these criteria, we collected 30 theorems in analysis and algebra from graduate-level qualifying examinations (including past exams from UCLA~\cite{ucla_past_qualifying_exams} and CU Boulder~\cite{cu_boulder_preliminary_exams}). These benchmarks expose common formalization bottlenecks where purely automated provers fail end-to-end.

\subsection{Experimental Setup}

We next describe the human participants who answer the queries generated by \tool, our evaluation metrics, and the baseline provers.

\paragraph{Human-in-the-loop evaluators.}
Because \tool is an interactive framework, our evaluation involves human participants who answer the mathematical questions it generates. We recruited undergraduate and graduate mathematics students with background in analysis and algebra. In each session, participants were shown queries produced by \tool and rendered in ordinary mathematical notation (see \autoref{fig:example-task}). For each query, participants indicated whether they could establish the stated goal from the given assumptions (\textit{yes} or \textit{no}), and their response times were recorded automatically. This setup measures the mathematical reasoning effort required from the user without requiring familiarity with Lean syntax.

\begin{figure}[t]\floatsize
    \centering
    \begin{tcolorbox}[paperbox]
        \uline{\textbf{Assumptions:}}

        \begin{tabularx}{\linewidth}{@{}>{\hsize=1.1\hsize\raggedright\arraybackslash}X@{\hspace{0.6em}}>{\hsize=0.9\hsize\raggedright\arraybackslash}X@{}}
        $\bullet$~$S$ is a set of points in the plane, that is, $S \subseteq \mathbb{R}^2$. &
        $\bullet$~$S$ contains exactly 5 points, i.e., $|S| = 5$. \\[0.4ex]

        $\bullet$~Any 3 points in $S$ are not collinear. &
        $\bullet$~$p$ is a point in $S$, i.e., $p \in S$. \\
        \end{tabularx}

        \medskip
        \uline{\textbf{Goal:}} Prove that $S$ is a finite set.

        \uline{\textbf{Question}}: Are you able to solve the goal? (\textit{yes/no})
    \end{tcolorbox}
    \caption{An example task and question presented to the evaluators.}
    \label{fig:example-task}
\end{figure}

\paragraph{Evaluation metrics.}
We evaluate \tool using three metrics:
\begin{itemize}
    \item \textbf{Number of Problems Solved:} The number of benchmarks (out of 30) verified end-to-end in Lean~4.
    \item \textbf{Interaction Count:} The number of queries answered by the user to complete a benchmark.
    \item \textbf{Resolution Time:} The total human time spent answering queries to complete a proof.
\end{itemize}

\paragraph{Baselines.}
We compare \tool against five state-of-the-art LLMs and theorem provers:
\begin{itemize}
    \item \textit{DeepSeek-Prover-V2~\cite{deepseek2025prover}:} An open-weights model specialized for formal theorem proving in Lean~4; we evaluate V2 rather than its general-purpose successor V3 due to its Lean-specific training.
    \item \textit{ChatGPT 5.1 Extended Thinking~\cite{openai_chatgpt}:} A reasoning model for multi-step mathematics, which also serves as the initial sketch generator in \tool.
    \item \textit{ChatGPT 5.4~\cite{openai_chatgpt54}:} OpenAI's general-purpose reasoning model.
    \item \textit{ChatGPT Codex 5.3~\cite{openai_codex53}:} A model specialized for code generation and formal language synthesis.
    \item \textit{Claude Opus 4.6~\cite{anthropic_claude46}:} Anthropic's reasoning model for long-context deduction.
\end{itemize}

\subsection{Effectiveness in Solving Real-World Problems (RQ1)}

To answer RQ1, we compare \tool against the union of all five baselines across our 30 benchmarks. Specifically, we count a benchmark as solved by the baselines if \textit{any} of the five models completes it.\footnote{For the baselines, we credit a model with success even if it axiomatizes atomic operations or simplifies intermediate steps, provided the overarching logic holds.} In contrast, \tool must produce a complete Lean~4 proof that reaches the \texttt{QED} state without any remaining \texttt{sorry} placeholders or weakened goals.

\begin{table}[t]\floatsize
    \centering
    \caption{End-to-end problem resolution across the 30 benchmarks. Baseline scores report the union of successes across all five baseline models.}
    \begin{tabular}{lcc}
        \toprule
        \textbf{System} & \textbf{Problems Solved} & \textbf{Success Rate} \\
        \midrule
        Baselines (One-Shot) & 6 & 20.0\% \\
        Baselines (Iterative) & 9 & 30.0\% \\
        \midrule
        \textbf{\tool (Ours)} & \textbf{28} & \textbf{93.3\%} \\
        \bottomrule
    \end{tabular}
    \label{tbl:rq1-results}
\end{table}

\autoref{tbl:rq1-results} summarizes the results under two baseline settings: \emph{One-Shot} (single-turn proof generation) and \emph{Iterative} (a repair loop where the model repeatedly revises its proof using compiler feedback from Lean~4).\footnote{A 20-minute timeout is enforced for iterative runs, whereas one-shot runs have no strict time limit.} The key takeaway from this experiment is that \tool solves 28 of the 30 benchmarks (93.3\%), whereas the union of all five baselines solves only 6 benchmarks (20.0\%) in One-Shot mode and 9 benchmarks (30.0\%) in Iterative mode.

To gain some intuition about these results, note that Mathlib often lacks high-level tactic automation for domain-specific bridge steps in graduate-level theorems. Consequently, purely automated provers must synthesize long sequences of low-level tactics on the fly, causing both single-turn generation and iterative compiler-guided repair to stall on unverified side conditions. In contrast, \tool isolates failing proof fragments via angelic checking and resolves their root causes through a small number of targeted queries to the mathematician.

Notably, \tool achieves this 93.3\% success rate using ChatGPT~5.1 Extended Thinking as its sketch generator, even though newer models such as ChatGPT~5.4 and Claude Opus~4.6 solve fewer problems when run standalone. This contrast suggests that the performance gains stem primarily from our refinement workflow rather than the raw capability of the underlying LLM: with symbolic residual analysis and targeted human judgment, the model does not need to synthesize every low-level tactic sequence from scratch.

\begin{mdframed}[
    backgroundcolor=gray!10,
    linecolor=white,
    innertopmargin=0.5em,
    innerbottommargin=0.5em,
    ]\noindent
    \textbf{Result for RQ1:} \tool solves 28 of the 30 benchmarks (93.3\%), whereas the union of five state-of-the-art automated provers solves 6 benchmarks (20.0\%) in One-Shot mode and 9 (30.0\%) in Iterative mode.
\end{mdframed}

\subsection{User Effort and Interaction Cost (RQ2)}

To answer RQ2, we evaluate the human effort required to complete proofs in \tool in terms of interaction count and resolution time, examine how participants with different mathematical backgrounds perform, and present a representative case study.

\paragraph{Aggregate interaction and time overhead.}
\autoref{tbl:rq2-overhead} reports the minimum, median, average, and maximum number of user interactions and total human resolution time across the 30 benchmarks.

\begin{table}[t]\floatsize
    \centering
    \caption{Aggregate human-in-the-loop overhead required to resolve the 30 benchmarks.}
    \begin{tabular}{lcccc}
        \toprule
        \textbf{Overhead Metric} & \textbf{Min} & \textbf{Median} & \textbf{Average} & \textbf{Max} \\
        \midrule
        \textbf{Interactions per Problem} & 1 & 5 & 6.23 & 15 \\
        \textbf{Resolution Time (seconds)} & 25 & 184 & 191 & 380 \\
        \bottomrule
    \end{tabular}
    \label{tbl:rq2-overhead}
\end{table}

As shown in \autoref{tbl:rq2-overhead}, completing a benchmark requires a \textit{median of 5 interactions} (average: 6.23, range: 1--15) and a \textit{median resolution time of 184 seconds} (average: 191 seconds, maximum: 380 seconds). Thus, even on graduate-level theorems that stall automated provers, \tool reduces the missing proof steps to a handful of yes/no questions that participants resolve in roughly three minutes on average.

\paragraph{Accessibility across participant backgrounds.}
Our evaluator pool included both graduate and undergraduate mathematics students. Notably, even though the benchmark theorems are drawn from graduate-level qualifying exams, undergraduate participants were frequently able to answer the resulting queries. Because \tool presents each query as a self-contained mathematical claim in ordinary notation, answering it requires standard mathematical judgment rather than familiarity with Lean.

\paragraph{Qualitative case study.}
\autoref{fig:rq2-case-study} illustrates how \tool reduces a formalization-heavy theorem to an accessible mathematical question. This Abstract Algebra benchmark asks to prove that for any field $F$, the polynomial ring $F[t]$ has infinitely many prime ideals. Although the LLM sketch follows the standard Euclidean proof strategy, its Lean script fails on several low-level steps involving irreducible factors, non-divisibility, and principal ideals. Rather than exposing these tactic errors to the user, \tool preserves the Euclidean argument as an angelic proof and applies residual analysis to isolate the root obligation.

As shown in \autoref{fig:rq2-case-study}, the resulting query reduces to a standard undergraduate algebra fact: given a finite set of nonconstant polynomials $S$ and $f = \big(\prod_{p \in S} p\big) + 1$, no polynomial $p \in S$ divides $f$ (since $p$ dividing both the product and $f$ would imply $p \mid 1$, which is impossible for a nonconstant polynomial). Once the participant confirms this statement, \tool automatically reconstructs the remaining ideal-theoretic steps in Lean.

\begin{figure}[t]\floatsize
    \centering
    \begin{tcolorbox}[paperbox, title=\textbf{Case Study: From Formal Obligations to a Mathematical Question}]
        \textbf{[Original Problem]: Abstract Algebra}\\
        Let $F$ be a field. Show that the polynomial ring $F[t]$ has infinitely many prime ideals.
        \tcbline
        \textbf{[\tool Decomposition]: Isolating the Conceptual Root}\\
        The system preserves the Euclidean proof skeleton via angelic execution, bypassing fragmented tactic errors regarding irreducible factorization and ideal-theoretic bridges.
        \tcbline
        \textbf{[Representative Final Query]}\\
        \textit{Question:} ``Given a finite set of nonconstant polynomials $S$, and $f = \big(\prod_{p \in S} p\big) + 1$, is it true that no $p \in S$ divides $f$?'' (\textit{yes/no})
    \end{tcolorbox}
    \caption{A qualitative example showing how \tool transforms a formalization-heavy Abstract Algebra proof into a single algebraic question stated in ordinary mathematical language.}
    \label{fig:rq2-case-study}
\end{figure}

\begin{mdframed}[
    backgroundcolor=gray!10,
    linecolor=white,
    innertopmargin=0.5em,
    innerbottommargin=0.5em,
    ]\noindent
    \textbf{Result for RQ2:} \tool requires a median of 5 interactions (6.23 on average) and a median of 184 seconds of human effort per problem, corresponding to roughly three minutes of a mathematician's time.
\end{mdframed}

\subsection{Ablation Study on Residual Analysis (RQ3)}\label{subsec:ablation}

To answer RQ3, we evaluate the impact of residual analysis by comparing the full version of \tool (\emph{full}) against an ablated variant (\emph{naive}) that disables residual analysis and queries the user on every leaf placeholder directly.

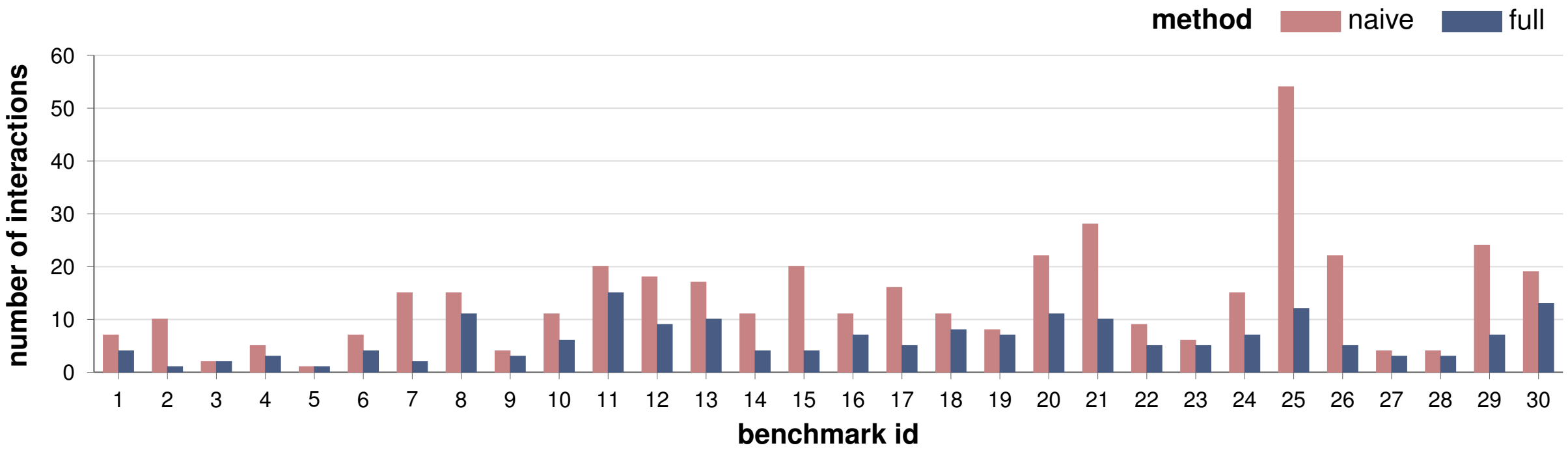
\begin{figure*}[t]
    \centering
    \begin{tikzpicture}
        \pgfplotstableread{
            x Naive Full
            1  7 4
            2  10 1
            3  2 2
            4  5 3
            5  1 1
            6  7 4
            7  15 2
            8  15 11
            9  4 3
            10 11 6
            11 20 15
            12 18 9
            13 17 10
            14 11 4
            15 20 4
            16 11 7
            17 16 5
            18 11 8
            19 8 7
            20 22 11
            21 28 10
            22 9 5
            23 6 5
            24 15 7
            25 54 12
            26 22 5
            27 4 3
            28 4 3
            29 24 7
            30 19 13
        }\rawdata

        \begin{axis}[
              ybar,
              bar width=4.2pt,
              area legend,
              width=\textwidth,
              height=4.8cm,
              tick label style={font=\sffamily\scriptsize},
              /pgf/number format/assume math mode=true,
              label style={font=\sffamily\footnotesize},
              bar shift auto=false,
              ymin=0, ymax=60,
              ytick={0,10,20,30,40,50,60},
              ylabel={\textbf{number of interactions}},
              xlabel={\textbf{benchmark id}},
              xlabel style={yshift=0.5ex},
              ylabel style={xshift=-0.05em},
              xmin=0.5, xmax=30.5,
              xtick={1,...,30},
              enlarge x limits=false,
              tick style={black!60},
              major tick length=2pt,
              legend style={
                  at={(1.0,1.02)},
                  anchor=south east,
                  draw=none,
                  fill=none,
                  font=\sffamily\footnotesize,
                  /tikz/every even column/.append style={column sep=0.6em},
              },
              legend columns=3,
              legend cell align=left,
              ymajorgrids=true,
              grid style={gray!25, line width=0.4pt},
              tick align=outside,
              tick pos=left,
              axis x line*=bottom,
              axis y line*=left,
              axis line style={black!60, line width=0.4pt},
          ]

        \addlegendimage{empty legend}
        \addplot+[fill=NaiveBarColor, draw=NaiveBarColor, bar shift=-2.2pt]
          table[x=x, y=Naive] {\rawdata};
        \addplot+[fill=FullBarColor, draw=FullBarColor, bar shift=2.2pt]
          table[x=x, y=Full] {\rawdata};

        \addlegendentry{\textbf{method}}
        \addlegendentry{naive}
        \addlegendentry{full}
        \end{axis}
    \end{tikzpicture}
    \caption{Ablation study across the 30 mathematical benchmarks. The chart compares the number of user interactions required per problem with and without residual analysis. Overall, residual analysis reduces total interactions by 55\%.}
    \label{fig:rq3-ablation-chart}
  \end{figure*}

\autoref{fig:rq3-ablation-chart} shows the number of interactions required per benchmark by the \emph{naive} and \emph{full} variants. Across all 30 benchmarks, disabling residual analysis increases the total number of queries from 187 to 416 (an average of 13.87 vs.\ 6.23 queries per problem). Thus, residual analysis reduces the total number of user interactions by \textit{55\%}. The reduction is particularly pronounced on proofs where a single missing lemma triggers a cascade of downstream failures: for instance, residual analysis reduces the query count from 54 to 12 on Benchmark~25 (a 78\% reduction) and from 10 to 1 on Benchmark~2 (a 90\% reduction). By isolating upstream root obligations in the proof dependence graph, residual analysis eliminates redundant low-level queries and focuses user interaction on a compact set of core mathematical steps.

\begin{mdframed}[
    backgroundcolor=gray!10,
    linecolor=white,
    innertopmargin=0.5em,
    innerbottommargin=0.5em,
    ]\noindent
    \textbf{Result for RQ3:} Residual analysis reduces the total number of user interactions across the 30 benchmarks by 55\% (from 416 to 187 queries), and by up to 90\% on individual problems.
\end{mdframed}

\subsection{Threats to Validity}

We conclude this section by discussing three main threats to the validity of our evaluation.

\paragraph{Reliability of participant responses.}
Human responses could vary with participant background or query phrasing. To minimize subjectivity, \tool restricts interactions to self-contained yes/no mathematical questions in standard notation and avoids exposing raw Lean syntax. We also conducted pilot trials prior to the main evaluation to ensure the consistency and clarity of the query phrasing.

\paragraph{Representativeness of benchmarks.}
Because human-in-the-loop evaluation limits the number of tasks that participants can inspect, our suite comprises 30 benchmarks. To keep the suite representative, we drew problems from graduate-level qualifying examinations spanning measure and integration theory, real and functional analysis, and group, ring, field, and polynomial theory.

\paragraph{Dependence on initial proof sketches.}
\tool relies on an upstream LLM to generate the initial proof sketch. If the initial sketch is entirely off-track, optimistic checking and local refinement cannot recover a valid proof. However, on 93.3\% of the benchmarks (RQ1), the LLM sketch could be refined by \tool into a complete Lean~4 proof.

\section{Related Work}\label{sec:related}

In this section, we discuss prior work that is most closely related to \tool across LLM-based theorem proving, interactive proof assistants, program synthesis, and learning-based proof search.

\paragraph{Large language models for autoformalization.}
Recent work has explored using large language models to translate informal mathematical problems into formal proof scripts~\cite{wu2022autoformalization}. While models such as Minerva~\cite{minerva2022} and Llemma~\cite{llemma} focus on informal mathematical reasoning, automated provers such as GPT-f~\cite{polu2022formal}, Thor~\cite{jiang2022thor}, and DeepSeek-Prover-V2~\cite{deepseek2025prover} combine language modeling with premise selection and tactic search~\cite{yang2019learning} to prove formal theorems. Despite strong results on competition-style problems, these provers frequently stall on multi-step graduate-level theorems when missing side conditions or library preconditions cause tactic search to diverge. \tool complements these provers by using optimistic checking to preserve the high-level structure of an imperfect LLM draft and resolving the remaining proof obligations with targeted human guidance.

\paragraph{Interactive and human-in-the-loop proof synthesis.}
Interactive theorem provers (ITPs) such as Coq~\cite{rocq,coqart}, Lean~\cite{lean}, and Isabelle~\cite{isabelle} rely on users to guide proof construction. Hybrid systems such as TacticToe~\cite{gauthier2021tactictoe} and GPT-f~\cite{polu2022formal} assist users by suggesting tactics at open goals, while tools such as Alectryon~\cite{alectryon2020} align formal proof states with natural-language explanations to improve readability. Unlike tactic-suggestion tools, which require users to interact directly with the prover's formal syntax at each failing step, \tool reasons globally over the proof dependence graph to identify a minimal-cost set of root obligations and presents its queries to the mathematician in ordinary mathematical notation.

\paragraph{Program synthesis, angelic execution, and abductive inference.}
Our approach draws inspiration from angelic execution~\cite{bodik2010angelic,das2015angelic} and counterexample-guided inductive synthesis (CEGIS)~\cite{solarlezama2006}, which evaluate partial programs under optimistic assumptions to isolate unknown specifications or prune infeasible candidates. Related work on abstraction-guided synthesis, including STITCH~\cite{bowers2023stitch} and SDCL~\cite{sdcl}, combines symbolic reasoning and abstract interpretation~\cite{cousot1977} to guide search, while abductive inference techniques~\cite{isil,mayur} compute minimal root-cause assumptions for verification failures with user interaction. \tool adapts these ideas to formal theorem proving: it evaluates incomplete Lean scripts under angelic semantics by replacing failing fragments with \icode{sorry} placeholders, and casts minimal residual core selection as a MaxSAT problem over the proof dependence graph to minimize user queries.

\paragraph{Continuous learning for formal methods.}
Systems such as HOList~\cite{bansal2019holist}, TacticZero~\cite{wu2021tacticzero}, and GPT-f~\cite{polu2022formal} use reinforcement learning and iterative training loops to improve tactic prediction over time. In contrast, \tool operates purely at inference time without fine-tuning the underlying LLM, treating the model as a black-box sketch generator and combining symbolic residual analysis with a small number of mathematical judgments to complete proofs. Using the recorded human judgments as training data to improve upstream proof models is a promising direction for future work.

\section{Conclusion and Discussion}\label{sec:concl}

This paper presented \tool, a framework that converts imperfect LLM-generated Lean proof drafts into fully verified proofs with minimal human interaction. At the core of \tool is angelic proof checking, which isolates failing proof fragments as symbolic placeholders while preserving the high-level structure of the LLM's argument. To minimize user effort, \tool constructs a proof dependence graph augmented with inferred mathematical relations and formulates residual analysis as a MaxSAT optimization problem to select a minimal-cost set of essential obligations. By querying the user only on this minimal residual core, \tool automatically reconstructs the complete Lean proof. Our evaluation on 30 challenging benchmarks across analysis and algebra shows that \tool solves 93.3\% of the problems while reducing the number of user interactions by 55\%.

Recent discussions of AI-generated mathematics have emphasized that human understanding remains essential alongside formal verification. Even when a proof has been checked by a proof assistant, mathematicians may still need substantial effort to identify its key ideas, understand the role of its assumptions, and relate the argument to existing theory. Interactive frameworks such as \tool offer a complementary mode of AI-assisted proving, in which mathematicians engage with selected intermediate claims while the proof is being refined, rather than only after it is complete. In \tool, these claims are presented in ordinary mathematical language, so that local mathematical judgments become part of the refinement process.

The implications of this interaction for mathematical understanding remain to be assessed. The current cost model prioritizes estimated verification effort and interaction count, without explicitly accounting for the explanatory role of the selected obligations. Moreover, yes/no judgments on individual statements do not by themselves establish understanding of the overall argument, and our experiments measure neither proof comprehension nor the effort required to study the completed proof. Future work could incorporate the explanatory role of obligations into query selection and present each query within a readable proof outline, showing why it arises and how it contributes to the argument. Further studies could assess whether such interaction improves proof comprehension and reduces the effort needed to study and reuse the resulting proofs.

\section{Declarations}

\paragraph{Conflicts of interest.}
The authors declare no conflicts of interest.

\paragraph{Use of AI-assisted tools.}
During the preparation of this manuscript, the authors used Anthropic Claude~5.1 and Google Gemini~4 (Argon) to assist with editing and polishing the wording of the text and with checking the consistency of notation and references. All AI-assisted content was reviewed, verified, and revised as necessary by the authors. ChatGPT~5.1 Extended Thinking is used as the proof-sketch generator within \tool; this use is part of the method and is described in \autoref{sec:impl} and \autoref{sec:eval}.

\paragraph{Acknowledgements.}
We thank the undergraduate and graduate mathematics students who participated in the evaluation sessions.

\bibliographystyle{unsrt}
\bibliography{ref}

\end{document}